\documentclass{article}

\usepackage{PRIMEarxiv}

\usepackage[utf8]{inputenc} 
\usepackage[T1]{fontenc}    
\usepackage{hyperref}       
\usepackage{url}            
\usepackage{booktabs}       
\usepackage{amsfonts}       
\usepackage{nicefrac}       
\usepackage{microtype}      
\usepackage{lipsum}
\usepackage{fancyhdr}       
\usepackage{graphicx}       
\graphicspath{{media/}}     
\usepackage{amsmath}
\usepackage{svg}
\usepackage{multirow} 
\usepackage{hyperref}  
\usepackage{doi}       
\usepackage{svg}       
\usepackage{placeins}  
\usepackage{float}      

\title{Large Language Models Acing Chartered Accountancy
\thanks{\textit{\underline{Citation}}: J. Gupta, A. Sharma, S. Singhania, M. Adnan, S. Deo, A. I. Abidi and K. Gupta. MoStart 2025: International Conference on Digital Transformation in Education and Applications of Artificial Intelligence. 2025.}
}

\author{
  Jatin Gupta \\
  Dept. of Computer Science and Engineering \\
  Sharda University, Greater Noida, India \\
  \texttt{jatingupta261001@gmail.com}
  \And
  Akhil Sharma \\
  Dept. of Computer Science and Engineering \\
  Sharda University, Greater Noida, India \\
  \texttt{sharmaakhil944@gmail.com}
  \And
  Saransh Singhania \\
  Dept. of Computer Science and Engineering \\
  Sharda University, Greater Noida, India \\
  \texttt{saransh060123@gmail.com}
  \And
  Mohammad Adnan \\
  Dept. of Computer Science and Engineering \\
  Sharda University, Greater Noida, India \\
  \texttt{mohammadadnan9431@gmail.com}
  \And
  Sakshi Deo \\
  Dept. of Computer Science and Engineering \\
  Sharda University, Greater Noida, India \\
  \texttt{sakshideo05@gmail.com}
  \And
  Ali Imam Abidi\thanks{Corresponding author: \texttt{aliabidi4685@gmail.com}} \\
  Dept. of Computer Science and Engineering \\
  Sharda University, Greater Noida, India \\
  \texttt{aliabidi4685@gmail.com}
  \And
  Keshav Gupta \\
  Dept. of Computer Science and Engineering \\
  Sharda University, Greater Noida, India \\
  \texttt{keshav.gupta@sharda.ac.in}
}

\begin{document}
\maketitle

\begin{abstract}
Advanced intelligent systems, particularly Large Language Models (LLMs), are significantly reshaping financial practices through advancements in Natural Language Processing (NLP). However, the extent to which these models effectively capture and apply domain-specific financial knowledge remains uncertain. Addressing a critical gap in the expansive Indian financial context, this paper introduces CA-Ben, a Chartered Accountancy benchmark specifically designed to evaluate the financial, legal, and quantitative reasoning capabilities of LLMs. CA-Ben comprises structured question-answer datasets derived from the rigorous examinations conducted by the Institute of Chartered Accountants of India (ICAI), spanning foundational, intermediate, and advanced CA curriculum stages. Six prominent LLMs—GPT-4o, LLAMA 3.3 70B, LLAMA 3.1 405B, MISTRAL Large, Claude 3.5 Sonnet, and Microsoft Phi 4—were evaluated using standardized protocols. Results indicate variations in performance, with Claude 3.5 Sonnet and GPT-4o outperforming others, especially in conceptual and legal reasoning. Notable challenges emerged in numerical computations and legal interpretations. The findings emphasize the strengths and limitations of current LLMs, suggesting future improvements through hybrid reasoning and retrieval-augmented generation methods, particularly for quantitative analysis and accurate legal interpretation.
\end{abstract}

\section{Introduction}
In the intricate and dynamic financial world, the integration of advanced intelligent systems, fueled by tech innovation, has entered the spotlight to transform financial processes \cite{Nie2024}.Of these, Natural Language Processing (NLP) has evolved into a key area of expansion and gained momentum after being incorporated into the global financial sector \cite{RodriguezInserte2023}. Large Language Models (LLMs), for instance, have established their immense applicability, performing well on sentiment analysis, text summarization, question-answering (QA), and language translation tasks, all well-suited to the complexities of financial language \cite{Samson2024}. However, the extent to which LLMs, e.g., ChatGPT, learn and apply finance-specific information during training remains largely unknown \cite{guo}. Benchmarks serve the important function of systematically evaluating and quantifying such uncertainties, allowing useful insight into the strengths and weaknesses of the models for financial reasoning \cite{Chang2024}. Despite the sheer scale of the Indian financial sector, a dedicated benchmark for assessing LLMs within this domain remains absent. This paper seeks to address this gap by proposing CA-Ben (Chartered Accountancy Benchmark), a benchmark for evaluating the performance of various LLMs within the Indian financial sector.

\section{Related Work}
Evaluating LLMs involves several key benchmark tasks and datasets. The datasets are divided into multiple tasks across the finance sector such as sentiment analysis, text classification, QA, etc.

In sentiment analysis, Financial PhraseBank provides 4,845 financial news sentences with sentiment labels \cite{Malo2013} and FiQA-SA offers 1,173 posts with sentiment scores \cite{Maia2018}. For text classification, the Headline Dataset categorizes 11,412 news headlines into nine classes, for example, "price up" or "price down"  \cite{Sinha2021}. In the realm of named entity recognition, the FIN Dataset, extracted from eight U.S. SEC loan agreements, is employed for credit risk assessment \cite{Cesar2015}. QA tasks are addressed by FinQA, which contains 8,281 QA pairs from S\&P 500 annual reports \cite{Chen2021} with ConvFinQA extending the dataset through conversational data \cite{Chen2022}. Stock movement prediction leverages datasets such as StockNet, which fuses historical prices with Twitter posts for 88 stocks \cite{Xu2018}, alongside CIKM18 \cite{Wu2018} and BigData22 \cite{Soun2022} that adopt similar multi-source approaches. Text summarization research is facilitated by the ECTSum dataset, which is 2,425 document-summary pairs of Earnings Call Transcripts with Reuters-provided summaries \cite{Mukherjee2022}.

The FinBen benchmark continues from such projects by providing an end-to-end open-source evaluation framework which addresses 36 datasets on 24 distinct finance tasks, such as new evaluations of stock trading and Retrieval-Augmented Generation \cite{Xie2025}. The Chinese Financial Language Understanding Evaluation (CFLUE) benchmark also evaluates models over more than 38,000 multiple-choice questions and 16,000 application cases for a range of tasks from text classification to machine translation \cite{Zhu2024}.

\section{Benchmark Design and Structure}
The proposed benchmark, CA-Ben, consists of 14 QA papers in different subjects, marked in the Chartered Accountancy (CA) examinations held every year which are a series of stiff tests held by the Institute of Chartered Accountants of India (ICAI) to prepare candidates as Chartered Accountants \cite{Agrawal2010}. 
Papers are arranged in a structured manner across three different levels of CA exams—Foundation, Intermediate, and Final—based on the ascending difficulty of the CA course. Each level of the exam, represented as \( L_i \), comprises some certain papers, represented as \( P_j \). Each paper comprises separate objective QA pairs, represented as \( Q_{jk}^{(\text{Markdown})} \), within individual markdown-compatible text files for the sake of keeping the notation uncomplicated and legible for LLMs. The overall benchmark structure is merely represented by Equation~\eqref{eq:benchmark}.

\begin{equation}
    \text{CA-Ben} = \sum_{i=1}^{3} L_i \quad \rightarrow \quad \sum_{j=1}^{14} P_j \quad \rightarrow \quad \sum_{k=1}^{N_j} Q_{jk}^{(\text{Markdown})}
    \label{eq:benchmark}
\end{equation}

\begin{figure}[h]
    \centering
    \includegraphics[width=\linewidth]{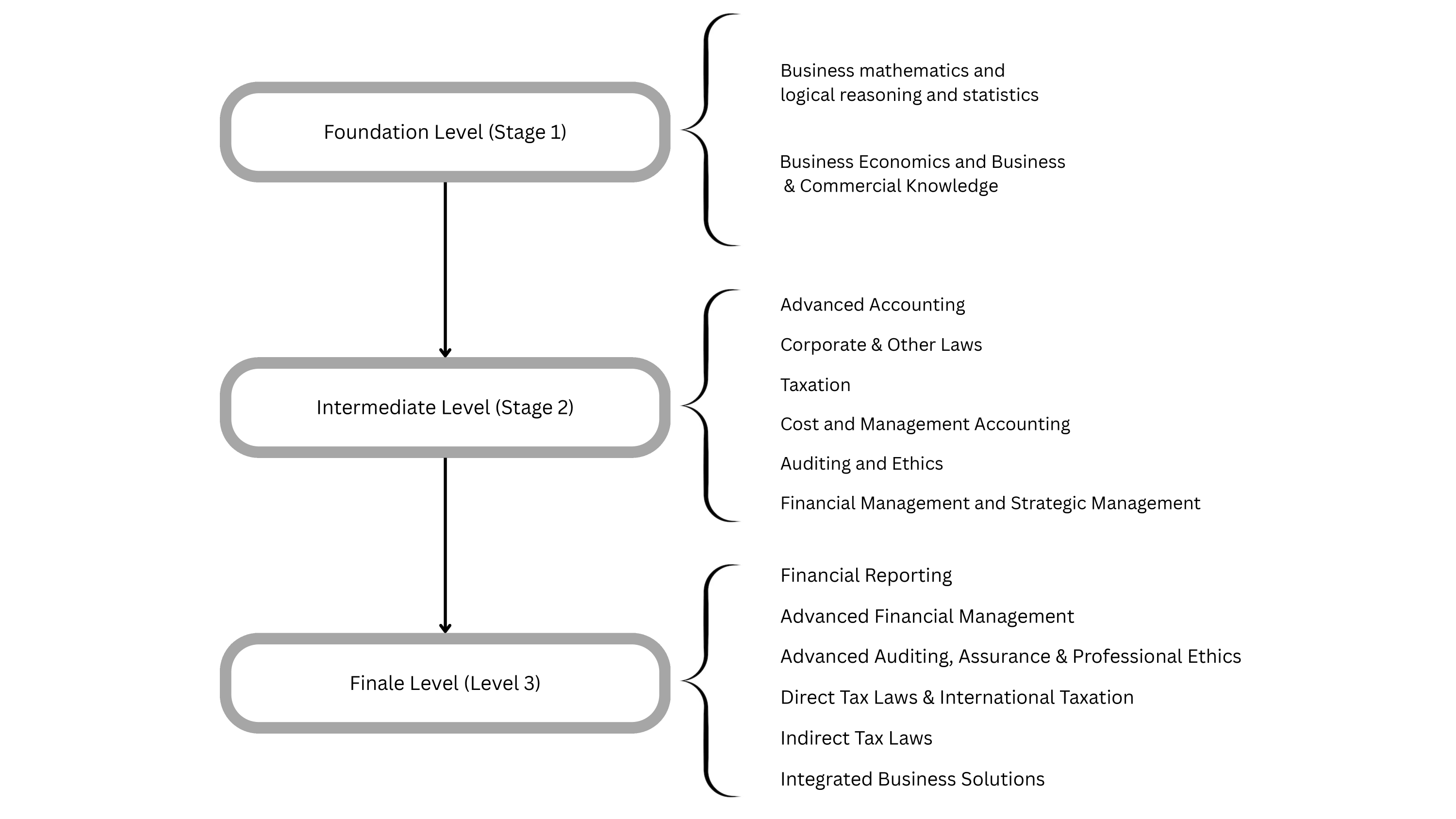}
    \caption{Hierarchical structure of the CA examination framework.}
    \label{fig:framework}
\end{figure}

The figure~\ref{fig:framework} shows a comprehensive diagram of the hierarchical structure of the CA examination format, indicating the different levels and subject categories in the Foundation, Intermediate, and Final levels.

At the CA Foundation level, emphasis is given to gaining basic knowledge in courses like Business Mathematics, Logical Reasoning, and Statistics, which enhance analytical abilities. Business Economics and Business and Commercial Knowledge also familiarize with basic economic principles and business activities.

CA Intermediate level is categorized into two groups. Group 1 includes Advanced Accounting, which strengthens the ability to prepare financial statements; Corporate and Other Laws, which include legal regimes that regulate business; and Taxation, which includes direct and indirect tax legislations. Group 2 includes Cost and Management Accounting, which involves cost control and decision-making processes; Auditing and Ethics, which focuses on auditing principles and professional ethics; and Financial Management and Strategic Management, which unifies financial planning with business strategy.

The CA Final level is the culmination of the examination process and is also split into two groups. Group 1 consists of Financial Reporting, which addresses complicated accounting standards; Advanced Financial Management, which addresses financial decision-making and investment planning; and Advanced Auditing, Assurance, and Professional Ethics, which addresses compliance with laws and upholds ethical auditing standards. Group 2 consists of Direct Tax Laws \& International Taxation and Indirect Tax Laws, which address complicated taxation systems, and Integrated Business Solutions, a culmination subject that integrates all previous knowledge to enhance strategic and analytical decision-making skills.

This benchmarking framework reflects the changing nature of the CA exam and thus guarantees that the dataset is a genuine representative of increasing complexity and knowledge range required at each level. The structured organisation of questions and their respective metadata enables effective benchmarking of LLMs in the field of professional accounting and financial reasoning.

\section{Implementation}
For the evaluation, six prominent state-of-the-art LLMs comprising of GPT-4o, LLAMA 3.3 70B Instruct, LLAMA 3.1 405B Instruct, MISTRAL Large, Claude 3.5 Sonnet, and Microsoft Phi 4, which were tested on the proposed benchmark. These models represent a diverse range of architectures and capabilities, varying in size, training data, and optimization techniques.

\subsection{System Prompt}
\begin{figure}[h]
    \centering
    \includegraphics[width=0.85\linewidth]{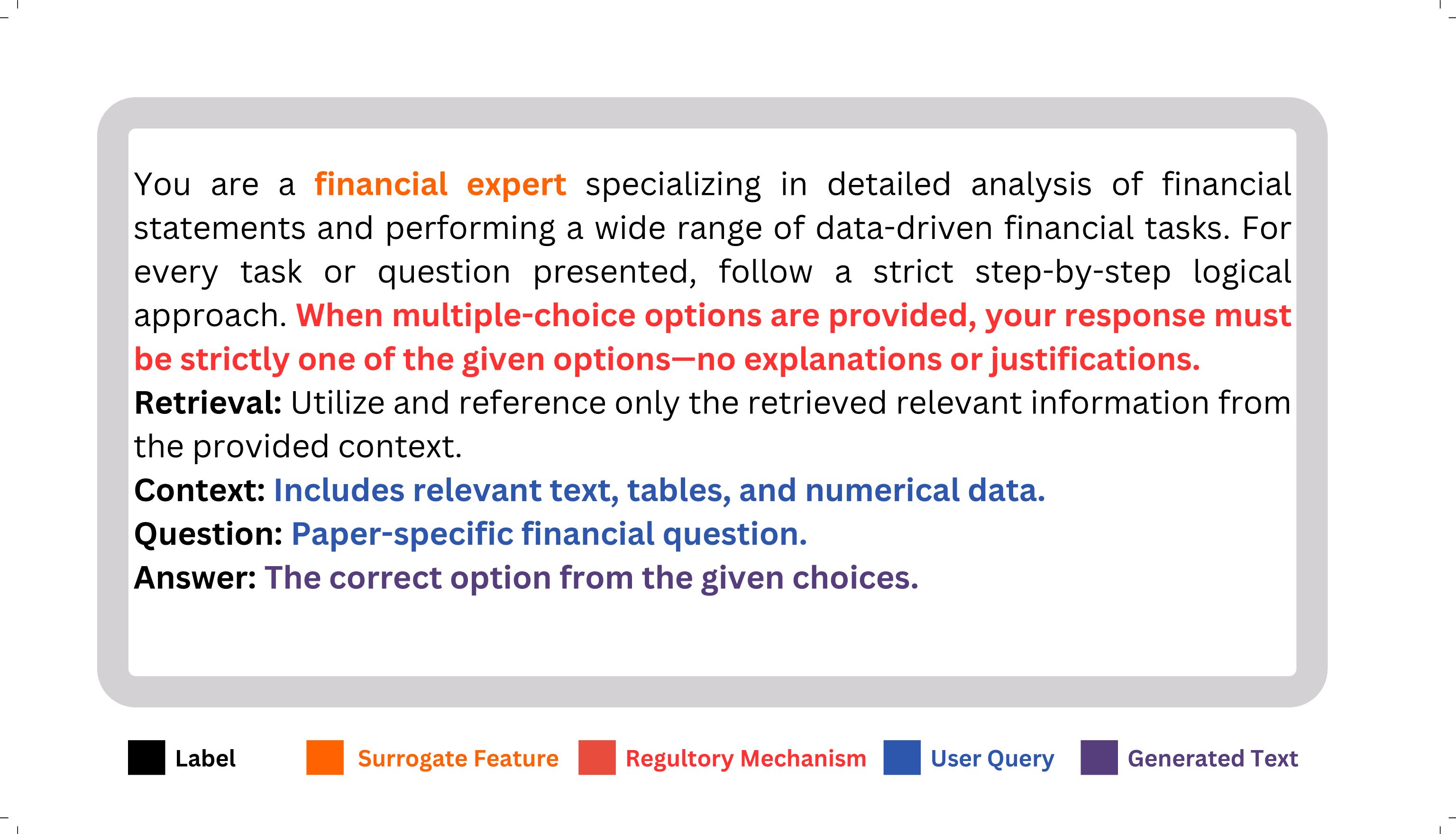}
    \caption{Structure of the System Prompt used.}
    \label{fig:prompt}
\end{figure}

A standardized system prompt was designed to ensure consistency in responses across all models. The system prompt is detailed in figure~\ref{fig:prompt}.

In the System Prompt, Label refers to a tag or category assigned to a piece of data, making it easier to classify or track. Surrogate Feature serves as a stand-in or proxy for an original variable when direct measurement is not possible or practical, ensuring that analysis can still proceed effectively. Regulatory Mechanism points to the rules or controls put in place—often for compliance or oversight—that directs how information is handled. User Query is simply the question posed by a user, prompting the need for analysis or a response. Finally, Generated Text represents the actual output produced in response to the User Query.

\subsection{Parameter Tuning}
A temperature setting of 0.75 was selected to introduce a balanced level of randomness in the responses, allowing the models to adapt to new scenarios and cases effectively. Lowering the temperature any further restricted the models' creative flexibility \cite{Peeperkorn2024} and it increased the likelihood of them not selecting any of the provided answer choices.

\subsection{Testing Pipeline}
The evaluation pipeline begins with the CA-Ben benchmark, where each QA pair \( Q_{jk} \), accompanied by a predefined System Prompt \( S_p \), is sequentially fed to six LLMs at temperature \( T = 0.75 \). The LLMs are denoted as LLM. This process is clearly depicted in (\ref{eq:pipeline}).

\begin{equation}
    P = \text{CA-Ben} \rightarrow (S_p + Q_{jk}) \rightarrow LLM(T = 0.75) \rightarrow R_m
    \label{eq:pipeline}
\end{equation}

The models’ responses \( R_m \) are then systematically parsed using a regex pattern to extract the final answer \( A_m \) choice (A-D) as shown in (\ref{eq:regex}).

\begin{equation}
    A_m = Regex(R_m)
    \label{eq:regex}
\end{equation}

where, 
\[
\text{Regex} = (?:\text{Result} | \text{Answer}):\\s*(\\?([A-D])\\)?
\]

The Regex pattern finds the nearest occurrence of the answer choice (A, B, C, or D) following the keywords "Result" or "Answer" in the generated text.

\begin{figure}[h]
    \centering
    \includegraphics[width=\linewidth]{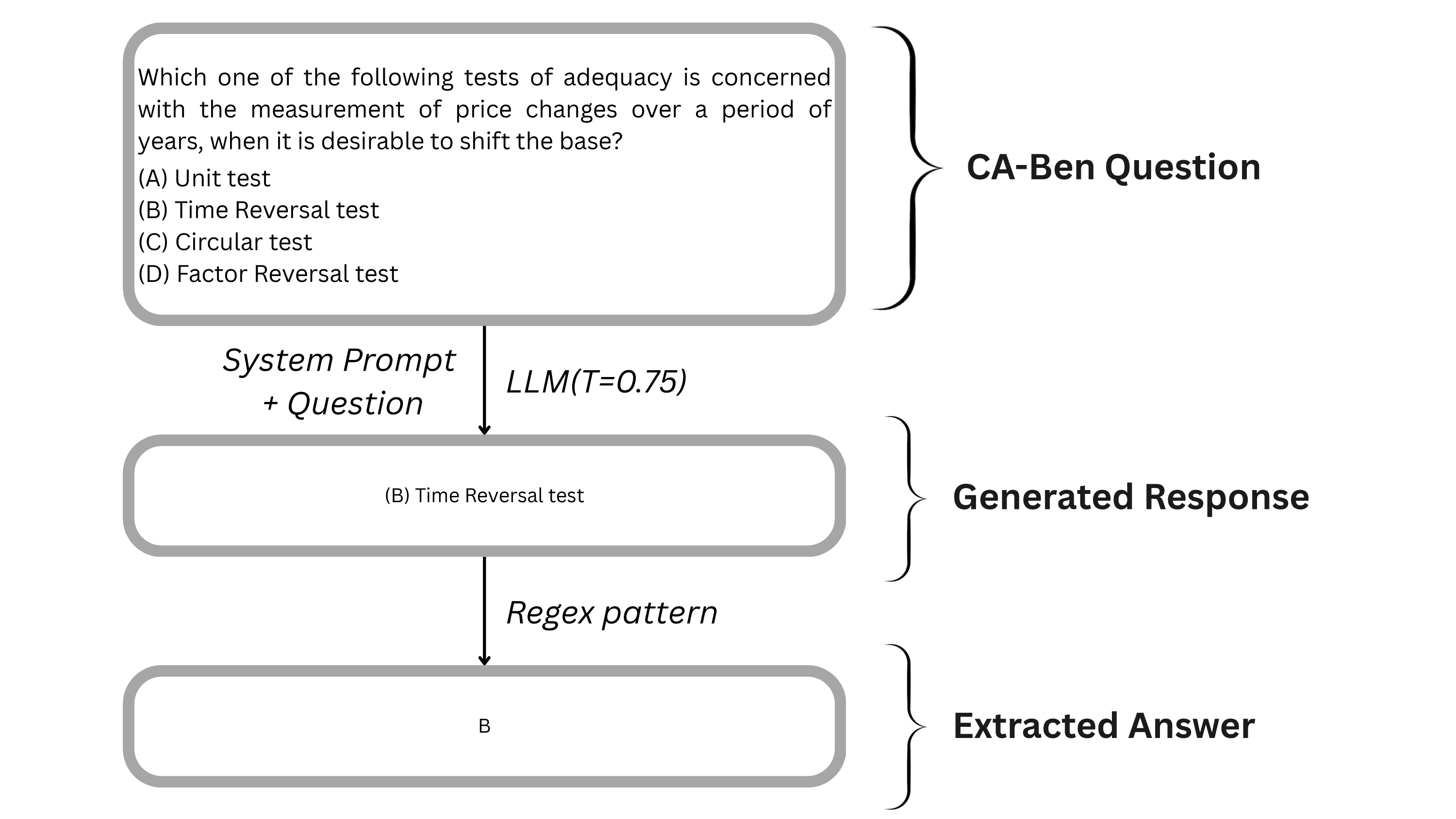}
    \caption{Example case of a query passed through the pipeline.}
    \label{fig:case}
\end{figure}

An example case of testing is illustrated in figure~\ref{fig:case}., where a Chartered Accountancy question from the CA-Ben dataset is combined with the standardized system prompt. This combined query is passed to the LLM at a temperature of 0.75, and the generated response is parsed to extract the selected answer for accuracy evaluation.

\section{Results and Evaluation}
\subsection{Benchmarking Criteria}
Accuracy was determined by calculating the percentage of correct answers extracted from the models' responses compared to the ground-truth labels. Formally, accuracy was computed using equation~\ref{eq:accuracy}.

\begin{equation}
    \text{Accuracy} = \frac{1}{N} \sum_{m=1}^{N} 1(A_m = G_m) \times 100
    \label{eq:accuracy}
\end{equation}

where \( A_m \) represents the extracted answer from each model’s response \( R_m \). \( G_m \) represents the ground-truth answer and \( N \) is the total number of attempted questions.

To align with the CA exam grading structure, two benchmarks were considered: an individual exam pass mark of 40\%, and a group-level qualification threshold of 50\%. These benchmarks provided additional context for evaluating model performance, simulating realistic examination criteria.

\subsection{Performance across Different Exam levels}
The results are categorized based on the three levels: Foundation, Intermediate, and Final. The cumulative accuracy of each model for these levels is presented in Table~\ref{tab:llm_accuracy}. For the complete results table of each subject, refer to \hyperref[sec:appendix]{Appendix A}.

\begin{table}[ht]
    \centering
    \caption{Overall Accuracy of LLMs Across Levels}
    \label{tab:llm_accuracy}
    \begin{tabular}{l@{\hspace{5mm}}c@{\hspace{8mm}}c@{\hspace{8mm}}c}

        \toprule
        \textbf{Models} & \textbf{Foundation} & \textbf{Intermediate} & \textbf{Final} \\
        \midrule
        GPT 4o & 54.00 & 53.61 & \textbf{55.28} \\
        LLAMA 3.3 70B Instruct & 57.5 & 53.61 & 41.65 \\
        LLAMA 3.1 405B Instruct & 56.00 & 49.37 & 44.01 \\
        MISTRAL Large & 48.5 & 46.59 & 38.47 \\
        Claude 3.5 Sonnet & \textbf{60.00} & \textbf{54.72} & 54.23 \\
        Microsoft Phi 4 & 59.00 & 49.23 & 41.63 \\
        \bottomrule
    \end{tabular}
\end{table}

At the Foundation Level, each model demonstrated robust performance, with Claude 3.5 Sonnet marginally leading ahead, closely followed by Microsoft Phi-4 and the remaining models. 
At the intermediate level, Claude 3.5 Sonnet continues to secure the highest cumulative score, with GPT-4o and LLAMA 3.3 70B Instruct following closely and demonstrating identical performance.
However, at the final level, overall performance across the models declined, with GPT-4o narrowly outperforming Claude 3.5 Sonnet.

\subsubsection{Performance on CA Foundation}
\begin{figure}[H]
    \centering
    \includegraphics[width=0.65\linewidth]{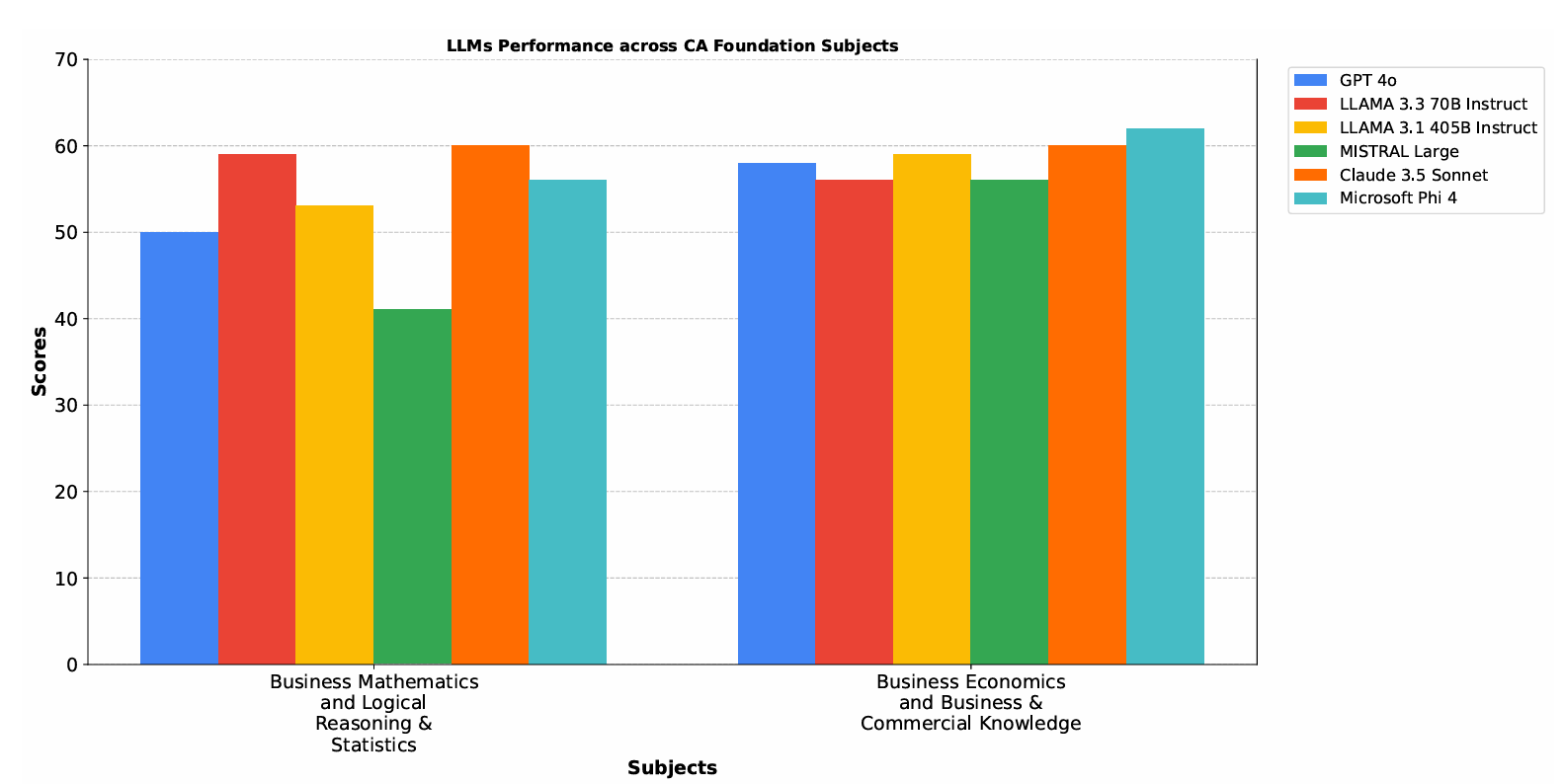}
    \caption{Accuracy of Models across Foundation Subjects.}
    \label{fig:img1}
\end{figure}

The models exhibited varying performance across subjects. Notably, Claude 3.5 Sonnet and LLAMA 3.3 70B Instruct demonstrated higher accuracy in subjects requiring logical reasoning and mathematical computation,  followed by the other models, as displayed in figure~\ref{fig:img1}.
Whilst in questions related to business, commerce, and Economics all models had nearly similar performance, with Microsoft Phi-4 outperforming the others marginally.

\subsubsection{Performance on CA Intermediate}
\begin{figure}[H]
    \centering
    \includegraphics[width=0.65\linewidth]{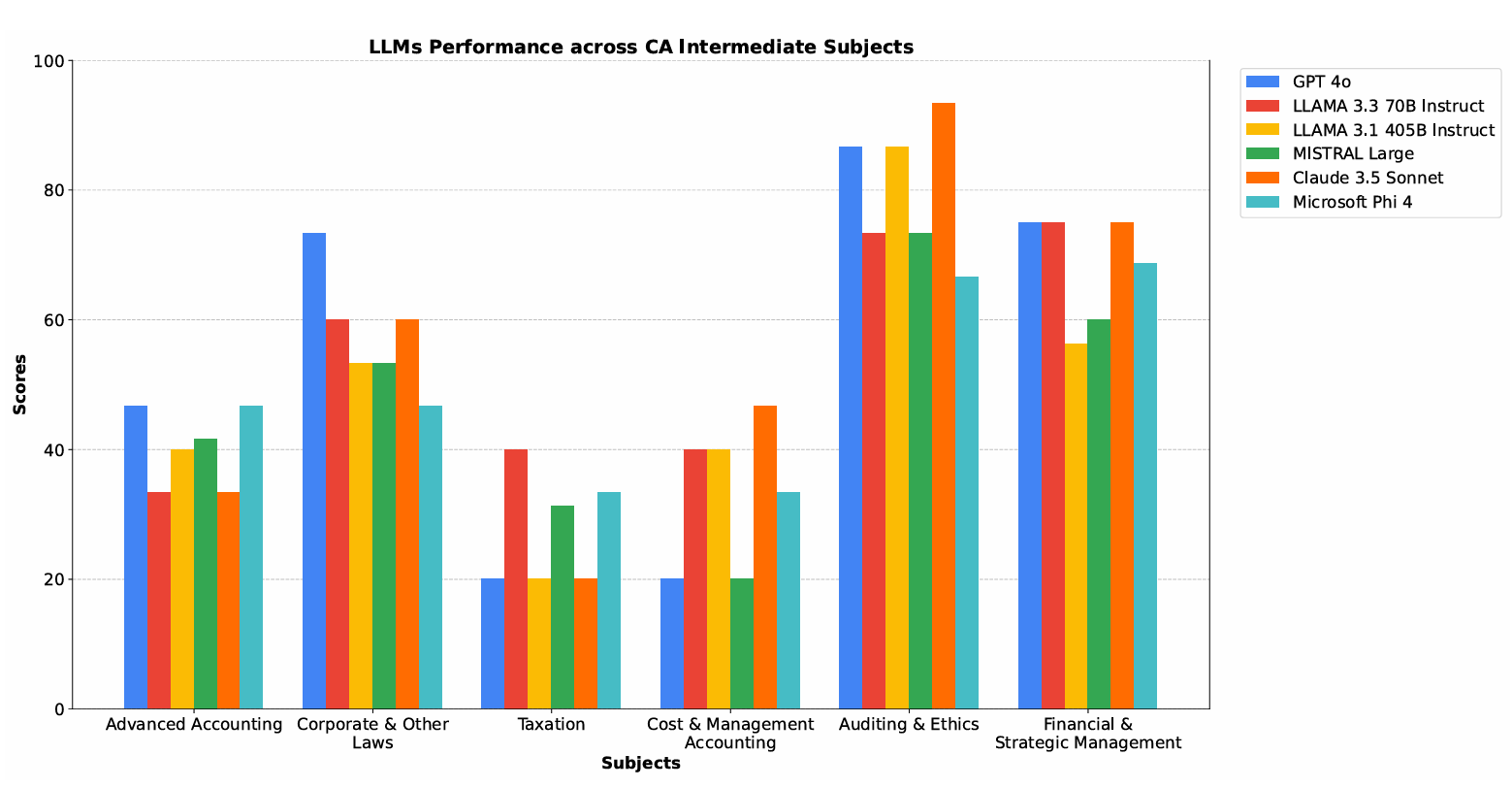}
    \caption{Accuracy of Models across Intermediate Subjects.}
    \label{fig:img2}
\end{figure}

Figure~\ref{fig:img2} illustrates the comparative accuracy of various LLMs across Intermediate-level CA subjects.
GPT-4o and Microsoft Phi-4 showed superior performance in Advance Accounting. 
In the domain of Corporate and Other Laws, GPT-4o exhibited superior performance, accompanied by LLAMA 3.3 70B Instruct and Claude 3.5 Sonnet, which shown equivalent efficacy.
In the sphere of taxation, LLAMA 3.3 70B outperformed the rest, with Microsoft Phi-4 standing at second. 
While for cost management and accounting, the Claude 3.5 Sonnet performed the best, followed by the two LLAMA models, which performed equally. Overall, the performance of all models in these two studies was not optimal. 
Contrary to the trend so far, for Auditing and ethics as well as financial and strategic management, the gross performance of all models was good, with both Claude 3.5 Sonnet and GPT 4o demonstrating excellence, with the LLAMA models following closely.

\subsubsection{Performance on CA Final}
\begin{figure}[H]
    \centering
    \includegraphics[width=0.65\linewidth]{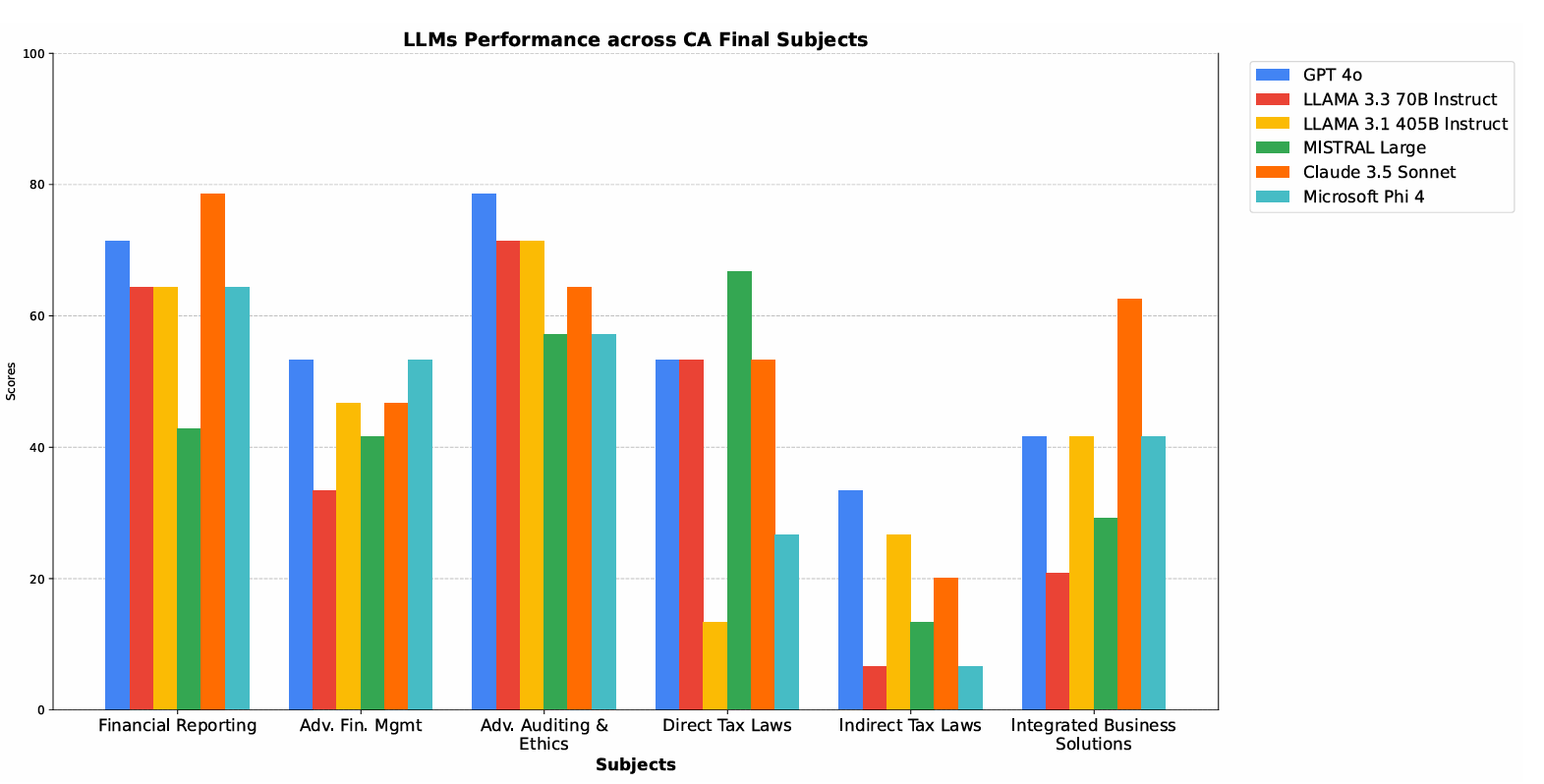}
    \caption{Accuracy of Models across Final Subjects.}
    \label{fig:img3}
\end{figure}

As shown in figure~\ref{fig:img3}, the comparative accuracy of various LLMs across Final-level CA subjects is presented.
In Financial Reporting, Claude 3.5 Sonnet scores the highest, followed by GPT-4o. 
For Advanced Financial Management, GPT-4o and Microsoft Phi-4 leads, while the LLAMA 3.1 405B and Claude models trail slightly behind.
In Advanced Auditing, Assurance, and Professional Ethics, GPT-4o dominates once more, followed by LLAMA models. 
In Direct Tax Laws and International Taxation, GPT 4o, LLAMA 3.3 70B Instruct and Claude 3.5 Sonnet are at the top, and LLAMA 3.1 405B at the bottom. 
For Indirect Tax Laws, GPT-4o again holds a clear lead, with LLAMA 3.1 405B Instruct and Claude 3.5 Sonnet near the next tier, and Microsoft Phi 4 and LLAMA 3.3 70B failing the test. 
Finally, in Integrated Business Solutions, Claude 3.5 Sonnet outperforms the others by a notable margin, followed by GPT 4.0, LLAMA 3.1 405B Instruct, and Microsoft Phi-4 clustered in the middle, and LLAMA 3.3 70B remains at the bottom.

\subsection{Model-Specific Comparative Analysis}
\begin{figure}[H]
    \centering
    \includegraphics[width=0.9\linewidth]{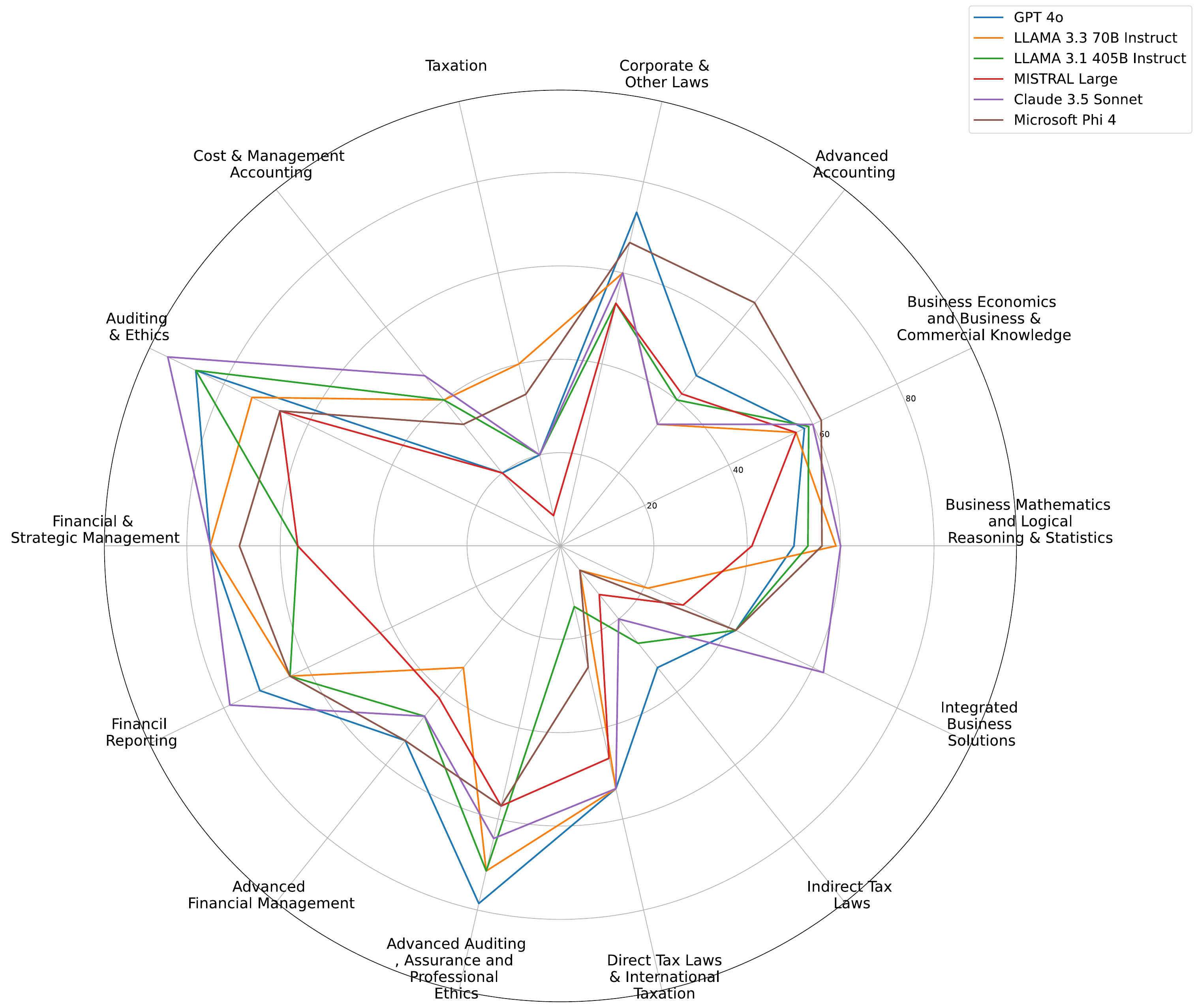}
    \caption{Model-specific comparative analysis: An overall comparison of models across each individual paper.}
    \label{fig:last}
\end{figure}

Each LLM demonstrated unique strengths across different subjects in the CA benchmark, reflecting their varied capabilities in handling financial, legal, and quantitative reasoning tasks. Figure~\ref{fig:last} clearly visualizes an overall comparison of models across each individual paper.

\subsubsection{Claude 3.5 Sonnet} emerged as the top-performing model with an average score of 56.31\%, showcasing exceptional results particularly in Auditing and Ethics (93.33\%), Financial Reporting (78.57\%), Integrated Business Solutions (62.50\%), and strong foundational understanding in Business Mathematics (60\%) and Business Economics (60\%), despite relatively weaker performance in Taxation (20\%) and Advanced Accounting (33.33\%).
\subsubsection{GPT-4o} with an average score of 54.29\%, displayed robust overall proficiency, excelling notably in Corporate and Other Laws (73.33\%), Advanced Auditing and Ethics (78.57\%), and Financial and Strategic Management (75\%), supplemented by commendable scores in Financial Reporting (71.43\%) and Direct Taxation (53.33\%). However, it exhibited modest performance in Taxation (20\%) and Cost \& Management Accounting (20\%).
\subsubsection{Microsoft Phi-4} achieved an average score of 49.95\%, reflecting moderate overall capability with notable strengths in Advanced Accounting (46.67\%), Business Economics (62\%), and consistent performance in Financial and Strategic Management (68.75\%) and Corporate Laws (46.67\%). Nevertheless, this model faced considerable difficulties in Indirect Taxation (6.67\%) and Direct Taxation (26.67\%), highlighting areas for substantial improvement.
\subsubsection{LLAMA 3.3 70B Instruct} scoring an average of 50.91\%, demonstrated fair competency primarily through Business Mathematics (59\%) and Financial and Strategic Management (75\%), complemented by adequate performances in Auditing and Ethics (73.33\%) and Corporate Laws (60\%). However, significant weaknesses emerged in Indirect Tax Laws (6.67\%), Integrated Business Solutions (20.83\%), and challenges persisted in Taxation (40\%).
\subsubsection{LLAMA 3.1 405B Instruct} delivered a modest overall average of 49.79\%, distinguished predominantly by strong results in Auditing and Ethics (86.67\%) and Advanced Auditing (71.43\%), along with reasonable foundational knowledge shown in Business Economics (59\%). However, the model struggled notably with Direct Tax Laws and International Taxation (13.33\%), Taxation (20\%), and underperformed significantly in Financial and Strategic Management (56.25\%).
\subsubsection{MISTRAL Large} was the weakest performer, averaging 44.52\%, exhibiting substantial limitations across most domains, notably poor in Taxation (31.25\%), Indirect Tax Laws (13.33\%), and Financial Reporting (42.86\%), despite moderate results in select areas like Business Economics (56\%) and Advanced Auditing (57.14\%), indicating a clear need for extensive refinement across the Chartered Accountancy curriculum.

\subsection{Limitations}
While LLMs demonstrated strong reasoning abilities, they faced notable challenges, particularly in numerical calculations, where multi-step computations often led to errors. Additionally, complex legal texts posed difficulties, with models sometimes misinterpreting jargon, resulting in incorrect answers. Hallucinations remains a challenge in giving null or void answers.

\section{Conclusion}
This study introduced CA-Ben, a comprehensive  financial benchmark for evaluating LLMs, within the Indian financial sector. By testing several LLMs—GPT-4o, LLAMA 3.3 70B Instruct, LLAMA 3.1 405B Instruct, MISTRAL Large, Claude 3.5 Sonnet, and Microsoft Phi-4—the study assessed their ability to reason through financial, legal, and mathematical concepts using the proposed benchmark. The results reveal variations in model performance, with some excelling in conceptual subjects like Business Economics, while others struggled with numerical accuracy and legal interpretation. Claude 3.5 Sonnet and GPT-4o achieved the highest scores, dominating the leaderboard in most papers and at various levels. These findings highlight both the potential and limitations of various LLMs in handling structured financial and regulatory content. 

\section{Future Scope}
Future advancements can enhance quantitative aptitude and mathematical accuracy by incorporating reasoning-based LLMs. A hybrid approach combining LLMs with symbolic reasoning, probabilistic frameworks and chain-of-thought prompting, can improve step-by-step problem-solving. Augmenting LLMs with legal terminology, via RAG (Retrieval-Augmented Generation) can significantly boost their capacity for accurate legal analysis and interpretation \cite{Gupta2024}.  Developing self-correcting AI architectures and neural-symbolic models will ensure precise calculations in taxation, financial modelling, and cost accounting, making AI more reliable for quantitative assessments in professional finance and accounting.

\bibliographystyle{unsrt}  
\bibliography{references}  

\clearpage  
\appendix
\section*{Appendix A: Consolidated Results of Model Evaluations}
\label{sec:appendix}

\subsection*{Foundations}
\begin{table}[h!]
\centering
\caption{Performance on Foundation-level Subjects}
\resizebox{\textwidth}{!}{
\begin{tabular}{|l|c|c|}
\hline
\textbf{Models} & \textbf{Business Math, Logical Reasoning \& Stats} & \textbf{Business Economics \& BCK} \\
\hline
GPT 4o & 50.00 & 58.00 \\
LLAMA 3.3 70B Instruct & 59.00 & 56.00 \\
LLAMA 3.1 405B Instruct & 53.00 & 59.00 \\
MISTRAL Large & 41.00 & 56.00 \\
Claude 3.5 Sonnet & 60.00 & 60.00 \\
Microsoft Phi 4 & 56.00 & 62.00 \\
\hline
\end{tabular}
}
\end{table}

\vspace{1em}  

\subsection*{Intermediate}
\begin{table}[h!]
\centering
\caption{Performance on Intermediate-level Subjects}
\resizebox{\textwidth}{!}{
\begin{tabular}{|l|c|c|c|c|c|c|}
\hline
\textbf{Models} & \textbf{Adv. Accounting} & \textbf{Corporate Laws} & \textbf{Taxation} & \textbf{Cost Mgmt Accounting} & \textbf{Auditing \& Ethics} & \textbf{Fin. \& Strategic Mgmt} \\
\hline
GPT 4o & 46.66 & 73.33 & 20.00 & 20.00 & 86.66 & 75.00 \\
LLAMA 3.3 70B Instruct & 33.33 & 60.00 & 40.00 & 40.00 & 73.33 & 75.00 \\
LLAMA 3.1 405B Instruct & 40.00 & 53.33 & 20.00 & 40.00 & 86.66 & 56.25 \\
MISTRAL Large & 41.66 & 53.33 & 31.25 & 20.00 & 73.33 & 60.00 \\
Claude 3.5 Sonnet & 33.33 & 60.00 & 20.00 & 46.66 & 93.33 & 75.00 \\
Microsoft Phi 4 & 46.66 & 46.66 & 33.33 & 33.33 & 66.66 & 68.75 \\
\hline
\end{tabular}
}
\end{table}

\vspace{1em}

\subsection*{Finals}
\begin{table}[h!]
\centering
\caption{Performance on Final-level Subjects}
\resizebox{\textwidth}{!}{
\begin{tabular}{|l|c|c|c|c|c|c|}
\hline
\textbf{Models} & \textbf{Financial Reporting} & \textbf{Adv. Fin. Mgmt} & \textbf{Adv. Auditing \& Ethics} & \textbf{Direct Tax Laws} & \textbf{Indirect Tax Laws} & \textbf{Integrated Business Solutions} \\
\hline
GPT 4o & 71.43 & 53.33 & 78.57 & 53.33 & 33.33 & 41.67 \\
LLAMA 3.3 70B Instruct & 64.29 & 33.33 & 71.43 & 53.33 & 6.67 & 20.83 \\
LLAMA 3.1 405B Instruct & 64.29 & 46.67 & 71.43 & 13.33 & 26.67 & 41.67 \\
MISTRAL Large & 42.86 & 41.67 & 57.14 & 46.67 & 13.33 & 29.17 \\
Claude 3.5 Sonnet & 78.57 & 46.67 & 64.29 & 53.33 & 20.00 & 62.50 \\
Microsoft Phi 4 & 64.29 & 53.33 & 57.14 & 26.67 & 6.67 & 41.67 \\
\hline
\end{tabular}
}
\end{table}

\end{document}